\documentclass[sigconf,authorversion,nonacm]{acmart}

\AtBeginDocument{%
  }

\usepackage{hyperref}
\usepackage{endnotes}
\usepackage{siunitx}
\usepackage{tabularx}
\usepackage{subcaption}
\begin{document}

\title{\textcolor{orange}{E2H}: A Two-Stage Non-Invasive Neural Signal Driven Humanoid Robotic Whole-Body Control Framework}
\author{%
  Yiqun Duan$^{1~\ast}$,
  Qiang Zhang$^{2~\ast}$,
  Jinzhao Zhou$^{1}$,
   Jingkai Sun$^{2}$,
  Xiaowei Jiang$^{1}$,
  Jiahang Cao$^{2}$,
  Jiaxu Wang$^{2}$,
  Yiqian Yang$^{1}$,
  Wen Zhao$^{3}$,
  Gang Han$^{3}$,
  Yijie Guo$^{3}$,
  Chin-Teng Lin$^{1}$
}

\affiliation{%
  \vspace{0.5em}
  \institution{$^{1}$Human-Centric AI Centre, AAII, University of Technology Sydney, Sydney, NSW, Australia\\
  $^{2}$The Hong Kong University of Science and Technology, Guangzhou, China\\
  $^{3}$Beijing Innovation Center of Humanoid Robotics, Beijing, China}
  \country{} 
}

\renewcommand{\shortauthors}{Yiqun et al.}

\begin{abstract}
Recent advancements in humanoid robotics, including the integration of hierarchical reinforcement learning-based control and the utilization of LLM planning, have significantly enhanced the ability of robots to perform complex tasks.  
In contrast to the highly developed humanoid robots, the human factors involved remain relatively unexplored.
Directly controlling humanoid robots with the brain has already appeared in many science fiction novels, such as \textbf{Pacific Rim} and \textbf{Gundam}.
In this work, we present \textcolor{orange}{\textbf{E2H} (\textbf{EEG-to-Humanoid})}, an innovative framework that pioneers the control of humanoid robots using high-frequency non-invasive neural signals.
As the none-invasive signal quality remains low in decoding precise spatial trajectory, we decompose the E2H framework in an innovative two-stage formation: 1) decoding neural signals (EEG) into semantic motion keywords, 2) utilizing LLM facilitated motion generation with a precise motion imitation control policy to realize humanoid robotics control. 
The method of directly driving robots with brainwave commands offers a novel approach to human-machine collaboration, especially in situations where verbal commands are impractical, such as in cases of speech impairments, space exploration, or underwater exploration, unlocking significant potential.
E2H offers an exciting glimpse into the future, holding immense potential for human-computer interaction.
\end{abstract}



\maketitle

\section{Introduction}

The public's attention to Humanoid robotics has rapidly risen, with significant contributions from leading companies and research institutes. Boston Dynamics’ \textit{Atlas} robot~\endnote{\href{https://bostondynamics.com/atlas/}{https://bostondynamics.com/atlas/}} showcases parkour-level mobility, while Tesla’s \textit{Optimus}~\cite{optimus} and \textit{Figure’s}~\endnote{\href{https://www.figure.ai/}{https://www.figure.ai/}} humanoids excel in complex manipulation tasks. 
\textit{Digit}\endnote{\href{https://agilityrobotics.com/robots}{https://agilityrobotics.com/robots}} and Unitree \textit{H1}\endnote{\href{https://www.unitree.com/h1/}{https://www.unitree.com/h1/}} also introduced their humanoid robot which can traverse diverse terrains with motor power.
Moreover, the utilization of LLMs also provides superior planning and embodied understanding ability for this humanoid robot, such as \textit{Figure}-2 and OpenAI’s \textit{1X} Robotics~\endnote{\href{https://openai.com/}{https://openai.com/}} did, highlighting the growing focus on machine intelligence itself. 

\begin{figure*}
    \centering
    \includegraphics[width=1\linewidth]{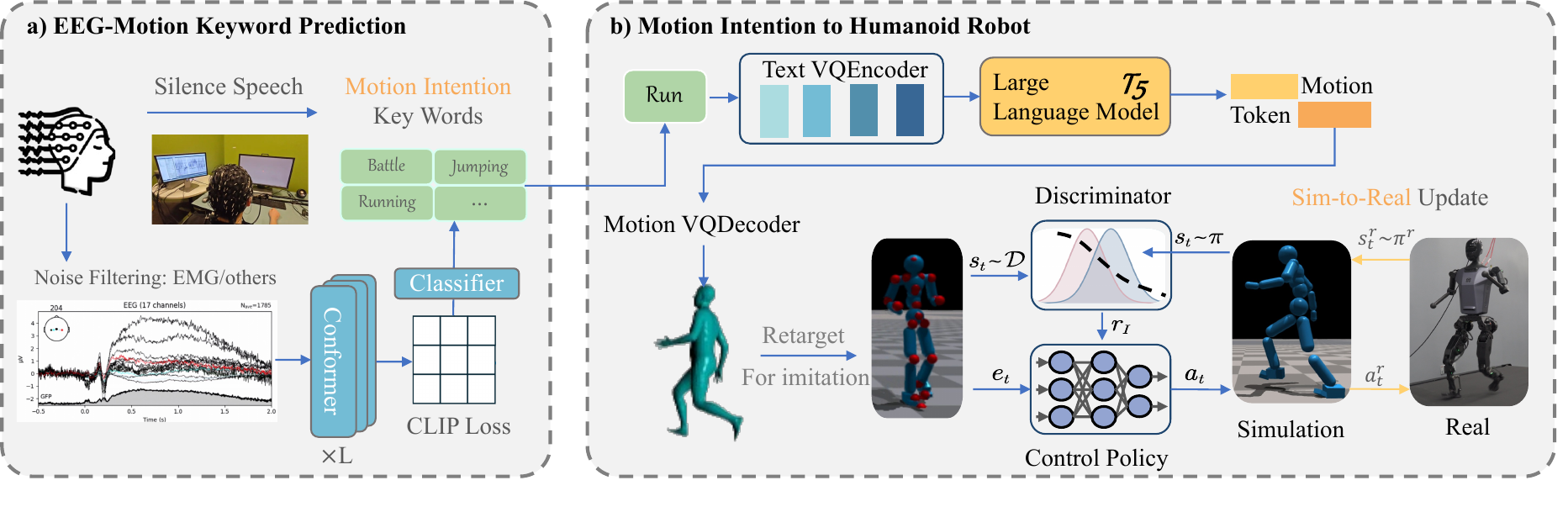}
    \caption{The illustration of the \textcolor{orange}{E2H} framework, which could be divided into two stages as described in a) and b). The conformed encoder $\&$ decoder structures shown in part a) first convert the human brain intention to discrete motion keywords. Then in part b) human intention to humanoid robotic control module converts the keyword motion intention to trajectory references. Based on the trajectory references. Given the generated motion reference, the controller model first learns to control policy in the simulation environment then the sim-to-real transfer is performed to get the control model for the physical robot. }
    \label{fig:mainflow}
\end{figure*}

Meanwhile, the interactions between humans and humanoid robots have become a prominent area of research~\cite{oztop2004human}. 
However, most of the previous work has focused on integrating human factors into humanoid robots by capturing human motion through poses. 
As early as 2002-2003, researchers at CMU~\cite{pollard2002adapting,safonova2003optimizing} were among the first to propose adapting captured human poses to control humanoid robots using traditional control algorithms. They even considered optimizing human motion to enhance the control of the robot. 
Following this direction, ASIMO~\cite{dariush2009online} introduced an improved control strategy.
With the recent prosperity of reinforcement learning in the field of humanoid robots and the advancement of simulation technology, H2O~\cite{he2024learning} proposed a motion-driven RL-based human-to-humanoid robots control framework. 

Considering only motion, there is no doubt that the human-computer interface lacks diversity. At the same time, such interaction completely restricts humans, who can only do synchronization but cannot do anything else. We boldly propose to directly use brain signals to control humanoid robots.
This new interaction opens up new possibilities, such as enabling more seamless teaming between humans and machines in various scenarios. For example, robots can collaborate with humans by following the intention when working together on construction sites or in rescue scenarios.
In particular, we propose to use non-invasive brainwaves (EEG), such an interaction does not rely on surgery and can be known to a wider user group.
Although the accuracy of translating EEG into motion intent is not very high at present, 
this paper still provides a feasible solution for this promising direction. 

Also due to the low data quality led by non-invasive data, direct decoding from the brain to precise control signal is not practical at this stage. 
To realize the interaction, we propose E2H and break the two-stage framework as illustrated in figure~\ref{fig:mainflow}. 
The first stage converted the EEG signal into motion keyword intentions (section~\ref{subsec:bcimodel}). Given the intention keywords, a moti
In the second stage, the robotic control system receives the motion intention and converts it into motion trajectories (section~\ref{subsec:trajectory}). Then the control policy controls the physical robot to fit the trajectory (section~\ref{subsec:control}).
To realize the critical brain intention to motion keywords translation, we design a data collection process through visual cues and let humans perform active actions (section~\ref{subsec:datacollection}). 
In the inference stage, we directly combine the two stages to realize brain-to-robotic control. 
The contribution of this work can be categorized into three parts. 
\begin{itemize}
    \item E2H is the first work that explores the direct bridge between humanoid robots and brain signal intention. By breaking down the framework into two stages, E2H could convert EEG signals into humanoid robot control signals, providing a novel way of interaction for human-humanoid robots teaming.
    \item We collected synchronized brain-to-motion keyword data across 10 different human subjects by designing an experimental pipeline in which participants performed active silent speech using 24 motion-related words to realize decoding of motion intention from EEG signals.
    \item To achieve more seamless motion control, we implemented a \textbf{neural feedback} mechanism where human subjects receive visual feedback from the robot during control attempts. If the robot fails to move, the subject can adjust its mental focus until the neural signals successfully drive the robot’s movement. Experimental results suggest successful control given multiple brain signals.
    
\end{itemize}
\section{Related Works}\label{sec:related}

\subsection{Learning Humanoid Robot from Human References}
Humans exhibit complex, versatile locomotion patterns that provide rich information for enhancing robot adaptability. Traditional behavior cloning methods, reliant on manual programming, were time-consuming and inflexible, making it challenging to define intricate humanoid robot skills~\cite{osa2018algorithmic,ravichandar2020recent,bohez2022imitate, han2023lifelike}. Recent advancements in imitation learning (IL) have shifted towards tracking reference joint trajectories or gait features~\cite{schaal1999imitation,van2010superhuman,jalali2019optimal,le2022survey}, though these methods often suffer from discontinuities during transitions between motion patterns. To address this, Peng et al. introduced Generative Adversarial Imitation Learning (GAIL) methods like AMP and Successor ASE, which allow physics-based avatars to perform tasks while imitating diverse motion styles from large datasets~\cite{ho2016generative,peng2021amp,peng2022ase}. These approaches have been applied to agile quadrupedal locomotion and terrain adaptation~\cite{escontrela2022adversarial, vollenweider2023advanced, li2023learning, wu2023learning, wang2023amp}, and were further refined by Tang et al. with a Wasserstein adversarial system~\cite{tang2023humanmimic}. Additionally, re-targeting techniques have been developed to transfer reference motions to robots while maintaining skeleton and geometry consistency~\cite{ayusawa2017motion,grandia2023doc,tang2023humanmimic,sun2024prompt,zhang2024whole}.

\subsection{Brain to Motion Intention Data Collection}
\label{subsec:datacollection}
\begin{figure*}
    \centering
    \includegraphics[width=0.9\linewidth]{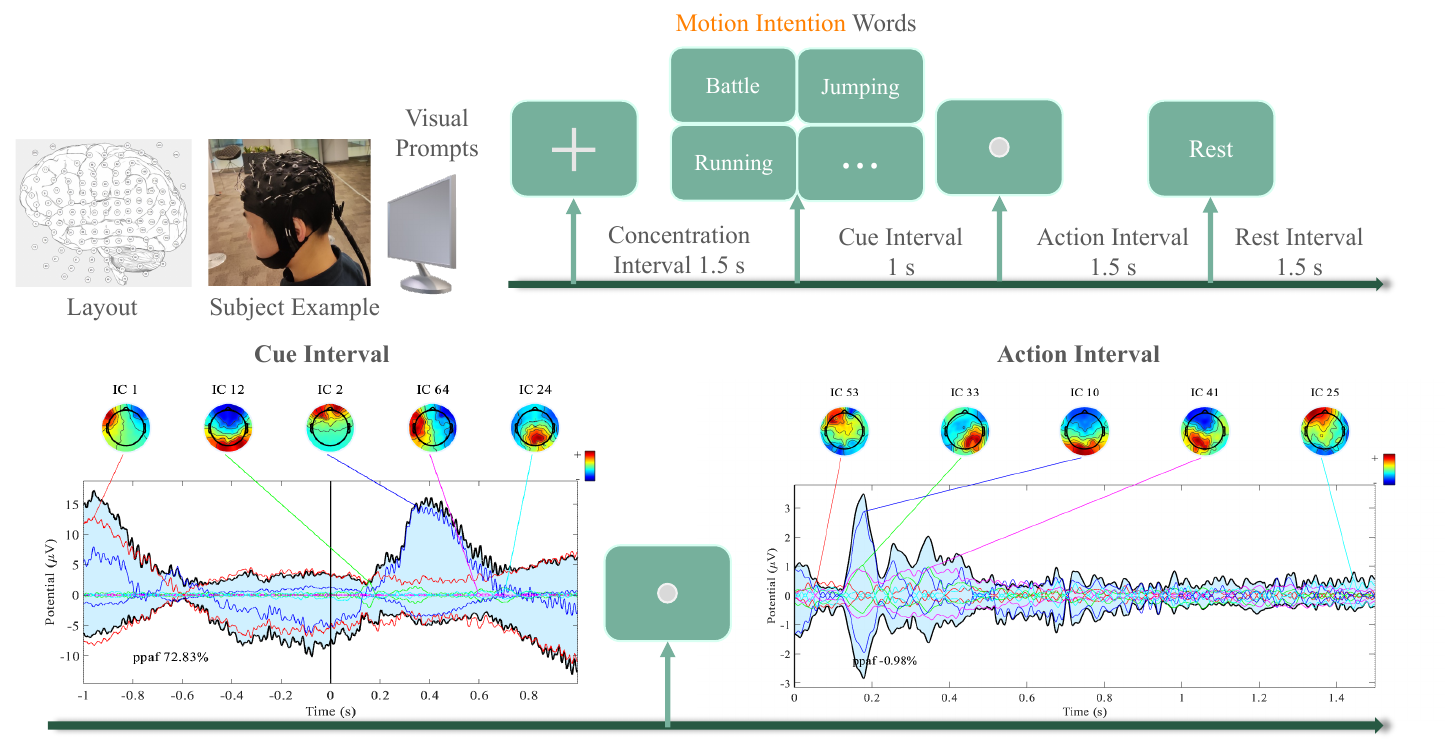}
    \caption{Illustration of the data collection process, where the left part illustrates the channel layout of the collection system. The EEG-to-motion synchronized data is collected by giving the human subjects a visual prompt as a cue and letting the subjects perform silent speech when actively thinking about the motion keywords. The lower part suggests the brain signal change during the data collection period.}
    \label{fig:datacollection_illustration}
\end{figure*}

\subsection{Brain to Semantic Decoding}

The field of converting brain signals into text has seen rapid advancements. In 2019, Anumanchipalli et al.\cite{anumanchipalli_2019_speech_synthesis_neural_spoken_sentences_ecog_bilstm_brain2speech} pioneered a model translating ECoG patterns into speech, igniting further research. Wang et al.\cite{wang_2020_stimulus_speech_ecog_decoding_cortex_gan_brain2speech} later used GANs to synthesize speech from ECoG data, while Willett et al.\cite{Willett_2021_ecog_brain2text_handwriting_rnn_lm_brain2text} developed an RNN system to decode handwritten letters from neural activity. Metzger et al.\cite{Metzger_2022_ecog_rnn_beam_search_gpt_brain2text} enhanced text decoding with GPT-2, and further expanded open-vocabulary decoding capabilities~\cite{Metzger_2023_ecog_hubert_birnn_brain2speech_brain2text_avatar}. Liu et al.\cite{Liu_2023_ecog_chinese_cnn_lstm_brain2speech_brain2pinyin} introduced a model decoding Chinese pinyin from ECoG signals, while Feng et al.\cite{feng2023high_seeg_chinese_brain2language} achieved text interpretation from SEEG recordings—these systems mostly rely on invasive methods. Non-invasive approaches include Meta’s contrastive learning-based brain-to-speech system with MEG and EEG data~\cite{D_fossez_2023_meg_eeg_clip_pretrain_meta_brain2speech}, though limited to fixed sentences, and Wang et al.\cite{wang2022open_aaai_eeg2text}, who used a pre-trained BART model to translate EEG features into text. Other models~\cite{duan2023dewave_brain2text,zhou2024masked,zhou2024towards,yang2024decode,yang2024mad} refined this with wave2vec, codex representations, and more powerful language model, yet these methods often depend on teacher-forcing, potentially inflating performance metrics without considering noise-injected inputs~\cite{jo2024are}.
\section{E2H: EEG to Humanoid Robot Control}
This section provides technical details of the E2H framework. Brain-to-motion synchronized data collection is first introduced in section~\ref{subsec:datacollection}. The brain signal decoding module is then introduced in section~\ref{subsec:bcimodel}.
The decoded motion intentions are further converted into motion trajectory references and utilized that trajectory to control the humanoid robotics as described in section~\ref{subsec:control}. 
However, it is noted that the brain signal may not always successfully drive the robot. In that case, section~\ref{subsec:joint} introduces the neural feedback mechanism in the joint inference phase.

The data collection involved a structured experimental design with 8 participants, each of whom underwent a few days of experiments while on each experiment day, the participant will go through 4 sessions. Each session began with a 10-second baseline recording where participants were instructed to relax and remain still. Following the baseline, 385 stimulation trials were presented. During each trial, participants were instructed to focus on a white cross displayed on the screen, which served as a fixation point. The trials began with a concentration interval of 1.5 seconds, after which the visual cue appeared. Each participant completed a total of 385 trials across the first two sessions, with the possibility of varying numbers of trials in the third session based on individual willingness and fatigue. The dataset aimed to capture the neural activity associated with different speech conditions, providing valuable insights into the brain's response during inner speech and its potential applications in BCI technology.

\begin{figure*}[t]
    \centering
    \includegraphics[width=0.75\linewidth]{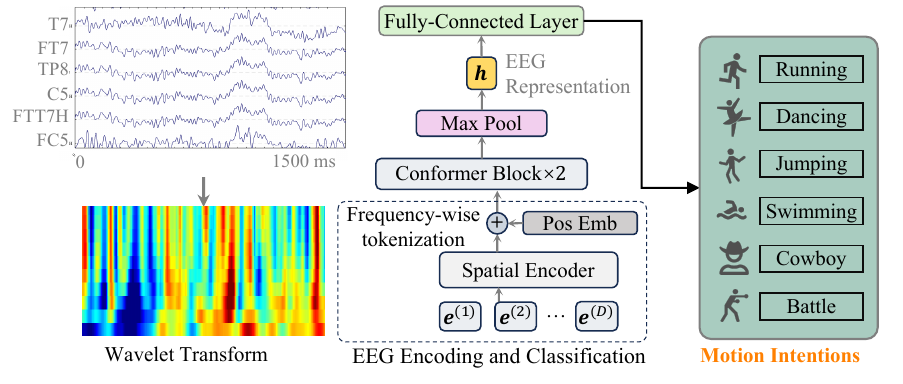}
    \caption{Illustration of the EEG classification model. In the classification task, we aim to predict the silently spoken class from the wavelet spectrogram of the multi-channel EEG signals using a Conformer encoder. }
    \label{fig:eeg-conformer}
\end{figure*}

\subsection{EEG signal to motion intention decoding.}
\label{subsec:bcimodel}


For perceiving users' motion intention directly from the brain signals, we collected an EEG dataset of 6 motion classes to train and evaluate the proposed EEG decoding model. In general, we have collected a total of 23.6 hours of EEG data covering 10 subjects for 6 motion types. The dataset has 128 channels with a sampling rate of 1000Hz. The EEG data has undergone several preprocessing steps to eliminate noise before using for training and evaluation. 1) We apply band-pass filters of \SI{0.5}{\Hz} to \SI{50}{\Hz} to the EEG signals to remove low-frequency drifting, line noise, and non-neural artifacts from the EEG signals. 2) CleanLine and average re-referencing operations are also used to improve the signal quality. 3) Afterward, we decompose the filtered EEG signal using independent component analysis (ICA) to identify high probability (>90$\%$) artifacts introduced by eye, muscle, and cardiac activities. 

After pre-processing, we used a wavelet filter to transform the time-series EEG wave data into wavelet frequency domain as primitive feature extraction for the EEG data before feeding it to the Conformer encoder. For encoding and classifying the EEG signals, we first perform tokenization on the frequency domain with $\mathbf{e}$ representing the multi-channel EEG frequency representation $\mathbf{e}\in\mathbb{R}^{N\times{D}}$. Here, $N$ denotes the number of channels, and $D$ is the number of frequency bands. To tokenzie the EEG, we split $\mathbf{e}$ into non-overlapping frequency bands across all channels $\{\mathbf{e}^{(i)}\}_{i=1,\cdots, D}$. Then, we tokenize the EEG frequency representation using a lightweight spatial encoder comprised of a convolutional layer and positional embedding for band location. The convolutional layer performs spatial filtering over the channel dimension while the positional embedding provides positional information about the frequency range.

We learn a conformer-based EEG encoder for encoding and classifying linguistic brain dynamics as illustrated in figure~\ref{fig:eeg-conformer}. The encoder outputs a probability distribution $p_{\theta}(\mathbf{y}|\mathbf{e})$, where we use $\theta$ to denote the parameters of the EEG encoder. After tokenizing the EEG frequency band information into latent space embeddings, we feed them to a Conformer model to build representations capturing global information across frequency bands. We train the output tokens of the Conformer model using both a self-supervised and supervised objective.

\subsection{Motion Intention to Humanoid Control}

\subsubsection{Humanoid Robot}

Researchers and society hold high expectations for Humanoids. Particularly, when combined with deep learning-based artificial intelligence, their potential becomes even greater~\cite{zhang2024whole}~\cite{wei2023learning}~\cite{zhang2024wococo}~\cite{fu2024humanplus}~\cite{cheng2024expressive}~\cite{zhuang2024humanoid}. Deep learning has achieved significant breakthroughs in visual and language tasks, and humanoid robots, due to their anthropomorphic form, possess unique advantages among all types of robots. This anthropomorphic characteristic enables humanoid robots to adapt to human living environments and directly utilize human data for imitation learning and training.

\begin{figure}[hbpt]
    \centering
    \includegraphics[width=1.0\linewidth]{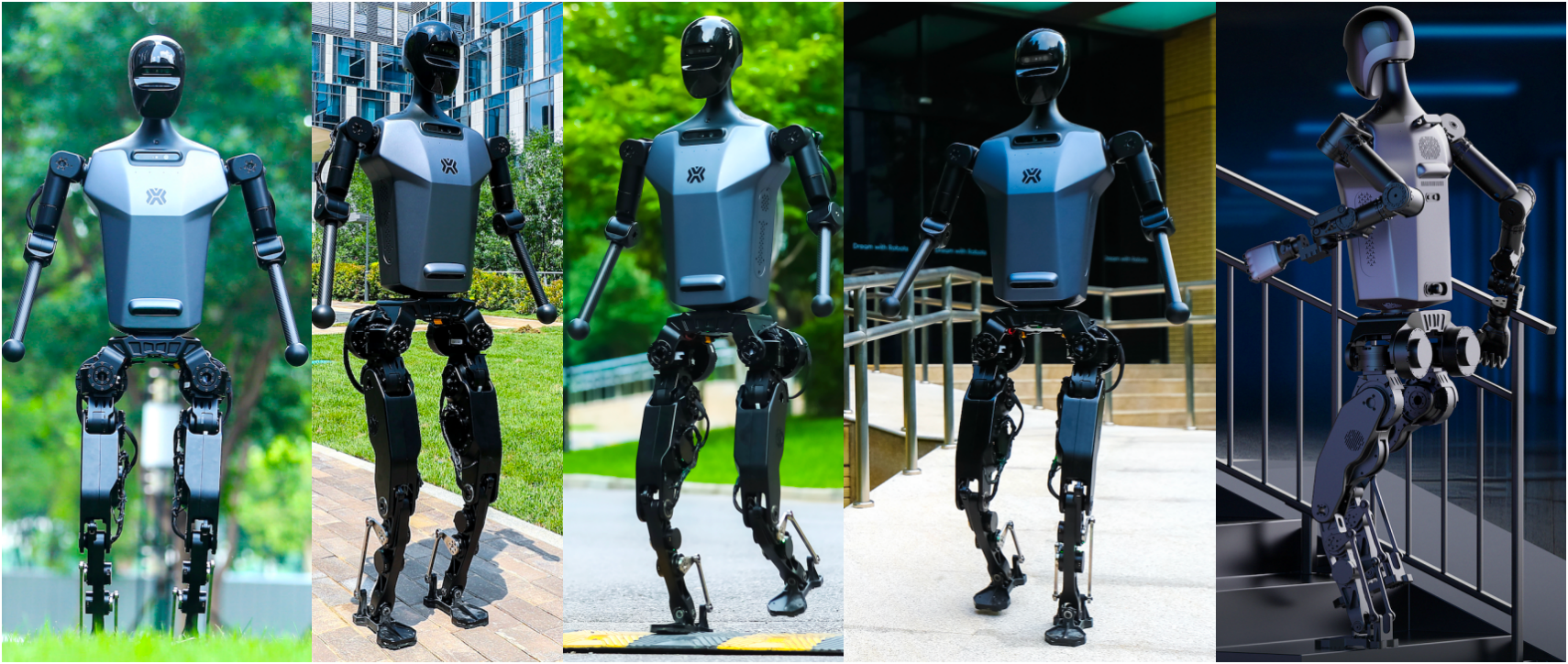}
    \caption{Illustration of the physical Tien Kung robots E2H utilized.}
    \label{fig:Presentation1}
\end{figure}

We present Tien Kung humanoid robots in Fig.\ref{fig:Presentation1}.
Which was developed by the Beijing Innovation Center of Humanoid
Robotics Co. Ltd. Tien Kung is a sophisticated humanoid
platform standing 163 cm tall and weighing 56 kg, equipped
with 42 degrees of freedom. Humanoid robots generally have limb structures that are very similar to humans, which allows them to adapt to human living environments and directly use human data for imitation learning. Regarding control, humanoid robots typically use remote controllers or visual feedback. However, due to the complexity of humanoid robots, language descriptions, and remote controllers often fail to precisely control such high complexity. While motion capture devices can provide precise control, they require very expensive manpower and specialized equipment. In this paper, we propose that EEG signals are a very promising and interesting control signal that can help humans interact with robots more naturally and intuitively.

\subsubsection{Convert Motion intention to Reference Trajectory}
\label{subsec:trajectory}
Once the abstract motion intention is decoded into text, we model the discrete text representations of the motion intention into continuous human motion. These concrete motion trajectories can serve as demonstrations to train humanoid whole-body control policy via imitation learning. Compared to abstract motion intentions, the text carries clear information, and the current advancements in large language models have equipped text with the powerful capability to transform it into other modalities. Therefore, we use the state-of-the-art Text-to-Motion model, named MotionGPT~\cite{jiang2024motiongpt}, to generate human motion data modeled by SMPL~\cite{loper2023smpl} from the text representing the motion intention. MotionGPT pre-trains a Vector Quantized Variational Autoencoders~(VQ-VAE) as motion tokenizer by following loss:
\begin{equation}
\begin{aligned}
    L_{vqvae} = \left\| \mathbf{x} - D(\mathbf{e}_k) \right\|_2^2 + \left\| \text{sg}[E(\mathbf{x})] - \mathbf{e}_k \right\|_2^2 + \beta \left\| E(\mathbf{x}) - \text{sg}[\mathbf{e}_k] \right\|_2^2
    \\
    \mathbf{z}_q(\mathbf{x}) = \text{Quantize}(E(\mathbf{x})) = \mathbf{e}_k \text{ where } k = \arg\min_i \left\| E(\mathbf{x}) - \mathbf{e}_i \right\|_2
\end{aligned}
\end{equation}
The motion tokenizer encodes the human motion as tokens, which is the same formulation as text. Subsequently, MotionGPT utilizes a unified text-motion vocabulary dataset to train a transformer-based model that addresses the conditioned generation task. It uses the log-likelihood of the data distribution as the loss function: 
\begin{equation}
    \mathcal{L}_{lm}=-\sum^{L-1}_{i=0}\text{log}~p(x_t^i|x_t^{<i},x_s)
\end{equation}
where $x_s$ is the input tokens sequence, $x_t$ is output token. The tokens can represent text tokens, motion tokens, or a mixture of two modalities. In our method, the input is text and the output is motion. Furthermore, to bridge the gap between human motion modeled by SMPL and the humanoid robots.

However, human motion lacks the physical, which leads to the robot performing impractical motions or even training failures. To address this problem, we utilize reinforcement learning~\cite{peng2021amp} to enable the network to interact with the environment. We formulate the training as a Markov Decision Process (MDP) with $(\mathcal{S},\mathcal{A},\mathcal{R},p,\gamma)$. $\mathcal{S}$ denotes the state space. $\mathcal{A}$ is the action space, $\mathcal{R}$ represents the reward function, $p$ means the transition probabilities from the current state $s_t$ to the subsequent state at next time step $s_{t+1}$, and $\gamma \in [0,1]$ is the discount factor. At each time step $t$, the agent outputs action depending on the state. Subsequently, the state of the robot transitions from $s_t$ to $s_{t+1}$ according to the transition function $s_{t+1} \sim p(s_{t+1}|s_t,a_t)$. The objective is to maximize the return reward by optimizing the parameters $\theta$ of the policy $\pi(a_t|s_t)$:
\begin{equation}
    \textnormal{arg}\max_{\theta} \mathbb{E}_{(s_t,a_t) \sim p_\theta(s_t,a_t)} \left[ \sum_{t=0}^{T-1} \gamma^t r_t\right]
\end{equation}
In the deployment of real humanoid robots, we need the end-effector position of humanoid limbs and the current desired foot contact states, which makes it hard 
to obtain SMPL-based visual motion. In the Initial phase, we follow the PHC~\cite{luo2023perpetual} to train a motion imitator to execute the reference motion in the physical-based simulator. The imitation policy receives the position and velocity of each human joint as the observation and tracks the corresponding motion. Due to PHC supervising the action in joint space, it is hard to apply to real humanoid robot deployment. Therefore, we convert the visual motion data into the robot data with dynamic constraints in this way. 
\subsubsection{Control}
\label{subsec:control}
To enable the robot to interact with the world stably and perform actions in the task space naturally. We introduce adversarial motion prior~\cite{peng2021amp} to force the policy to execute motion in human style rather than tracking joints of demonstration. AMP tackles some limitations of normal imitation learning such as behavior cloning by designing a discriminator $D(s_t,a_t)$ to measure the similarity between a policy and demonstrations. Then the reward of the discriminator is formulated as 
\begin{equation}
    r_I = \text{max}[0,1-\frac{1}{4}(D(s_t^{I}, s_{t+1}^{I})-1)^2]
\end{equation}
$s_t^I$ denotes the partial states applied to AMP. During each iteration, a randomly selected type of motion is input into the discriminator. In this setting, the agent is rewarded more favorably only if its executed actions closely align with the selected motion type. In addition to the traditional reinforcement learning loss, the loss for AMP is modeled as
\begin{equation}
\begin{aligned}
    L_{AMP} = &\frac{1}{2}\mathbb{E}_{(s_t^{I}, s_{t+1}^{I})\sim\mathcal{D}}[(D(s_t^{I}, s_{t+1}^{I})-1)^2] \\ +&\frac{1}{2}\mathbb{E}_{(s_t^{I}, s_{t+1}^{I})\sim\pi}[(D(s_t^{I}, s_{t+1}^{I})+1)^2] \\ +&\lambda_{GP}\mathbb{E}_{{(s_t^{I}, s_{t+1}^{I})\sim\mathcal{D}}}\left[\| \triangledown \mathcal{D}({s_t^{I}, s_{t+1}^{I}}) \|^2\right]
\end{aligned}
\end{equation}
where $(s_t^{I}, s_{t+1}^{I})\sim\mathcal{D}$ and $(s_t^{I},s_{t+1}^{I})\sim\pi$ mean the states transitions sampled from references or generated by the policy. In our humanoid whole-body control framework, we define the action space of RL policy as each desired joint angle $a_t \in \mathbb{R}^{20}$. The observation $o_t \in \mathbb{R}^{96}$ includes the average and current velocity of the root, position, velocity of each joint, and the orientation of the gravity vector in the robot's base frame. The desired position of humanoid limbs end-effector and foot contact states are $e_t$ added to the observation for precise motion tracking. 
\begin{table}[hbpt]
    \centering
    \caption{Regularization Rewards for Sim-to-Real Transfer}
    \scalebox{1}{
\begin{tabular}{ll}
\toprule
Reward Item                     & Formulation \\ \midrule
Action differential      & $\text{exp}(-\|\mathbf{a}_t - \mathbf{a}_{t-1}\|_2)$   \\
Joint velocity             & $\text{exp}(-\|\dot{\mathbf{q}}\|^2_2)$   \\
Joint acceleration         & $\text{exp}(-\|\ddot{\mathbf{q}}\|^2_2)$   \\
Torques                  & $\text{exp}(-\|\mathbf{\tau}\|_2)$    \\
\bottomrule
\end{tabular}
    }
    \vspace{-2mm}
    \label{tab: regularization rewards}
\end{table}

\begin{figure*}[hbpt]
    \centering
    \includegraphics[width=1\linewidth]{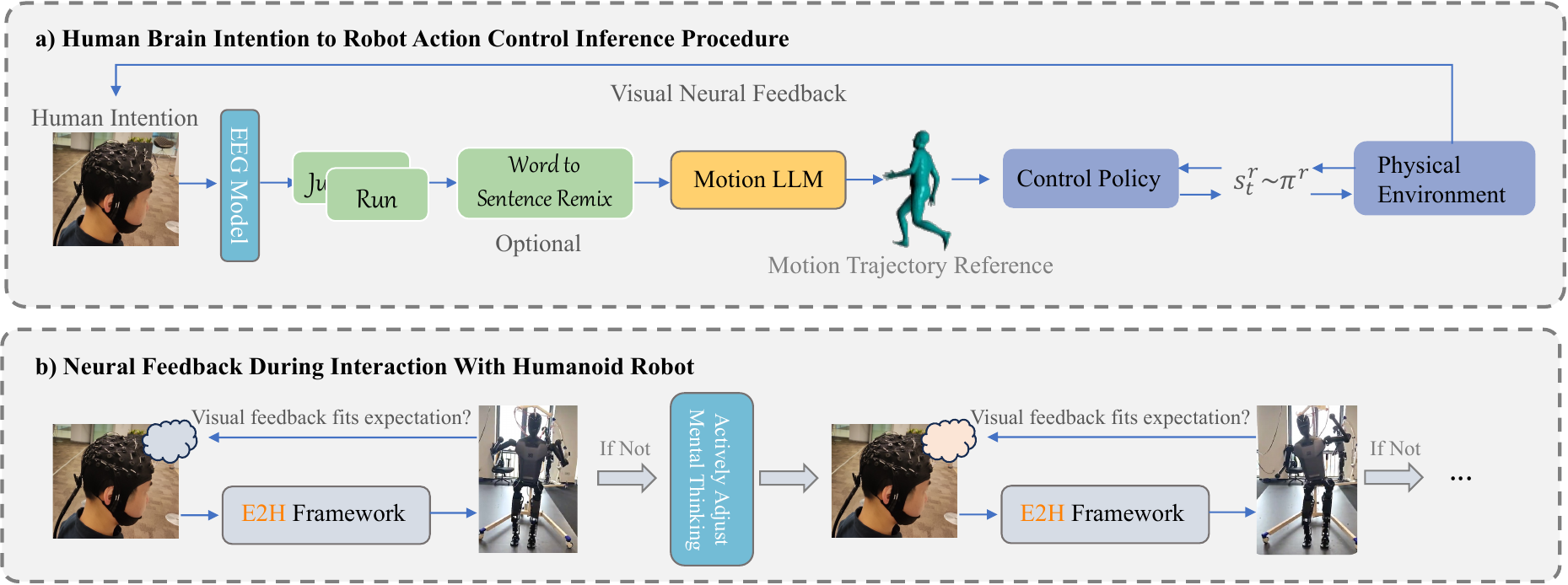}
    \caption{The inference flow of the proposed E2H framework. Subfigure a) denotes how the modules are connected during the inference stage. Subfigure b) denotes that during the interaction, the human subject observes the physical robot and actively adjusts its neural signal through visual feedback.}
    \label{fig:inference}
\end{figure*}
\begin{table*}[hbtp]
\centering
\caption{EEG to motion intention decoding performance. \label{tb:eegmodel}}
\begin{tabular}{c|c|c|c|c|c|c|c|c|c|c}
\toprule
\textbf{Subject} & \textbf{S05} & \textbf{S07} & \textbf{S08} & \textbf{S09} & \textbf{S10} & \textbf{S11} & \textbf{S12} & \textbf{S13} & \textbf{S14} & \textbf{S15} \\ \midrule
\multicolumn{11}{c}{\textbf{Accuracy@Top 1 (6 words)}} \\ \midrule
Conformer (\%)  & 42.76 & 33.08 & 29.67 & 42.96 & 33.54 & 33.54 & 40.74 & 42.44 & 54.47 & 31.61 \\ \midrule
\multicolumn{11}{c}{\textbf{Accuracy@Top 1 (24 words)}} \\ \midrule
Conformer (\%)  & 40.33 & 29.53 & 20.13 & 22.96 & 24.50 & 15.66 & 35.59 & 24.95 & 34.44 & 26.50 \\ \midrule
\multicolumn{11}{c}{\textbf{Accuracy@Top 3 (6 words)}} \\ \midrule
Conformer (\%)  & 76.31 & 59.55 & 55.48 & 57.03 & 66.45 & 65.80 & 68.88 & 64.02 & 82.08 & 62.58 \\ \midrule
\multicolumn{11}{c}{\textbf{Accuracy@Top 3 (24 words)}} \\ \midrule
Conformer (\%)  & 66.83 & 48.83 & 39.60 & 42.26 & 45.16 & 31.50 & 55.13 & 44.59 & 53.74 & 48.33 \\ \bottomrule
\end{tabular}
\label{tab:accuracy}
\end{table*}

\subsubsection{Sim-to-Real}
To improve the sim-to-real transfer, we incorporated the regularization rewards into the framework. The regularization rewards aim to reduce the disturbance caused by network output and improve smoothness and safety. The reward items are formulated as Table~\ref{tab: regularization rewards}. $\mathbf{a}_t$ denotes the action generated by policy, $\dot{q}$ and $\ddot{q}$ mean the velocity and acceleration of each joint. The action differential reward forces the network to output smoother action, which reduces the jitter of the whole-body humanoid robot. The rest of the regularization rewards respectively limit the velocity, acceleration, and torques of the robot to avoid motor overload. In the real-world deployment of humanoid robots, we use NVIDIA Jeston Orin to infer our model. We implement our approach on "Tien Kung" with 163cm tall and weighs 40kg. It has 20 degrees of freedom with whole-body QDD(quasi-direct drive) motors.

\subsection{Joint Brain-Robot Interaction}

Given the EEG-to-motion decoding module and the motion-to-robotic control module trained as described above, the whole pipeline is constructed as illustrated in figure~\ref{fig:inference}. 
In the beginning, the human subject will actively "think" the motion intentions by performing the same formation of "silent speech" without muscle movement. 
The EEG model receives the EEG data during this period and converts the neural signals into motion keywords. 
The motion keywords are further fed into a MotionLLM~\cite{jiang2024motiongpt} to get the reference motion trajectories.
Then the learned reinforcement learning-based controller controls the humanoid robot to fit the generated reference trajectories while adapting to different terrains and physical environments. 

\textbf{Neural Feedback} 
Because the non-invasive neural signal brings convenience but also lower data quality, the decoding of the motion intention is not always correct. According to the experiment section, the average accuracy of single-word motion intention decoding is around $40 \% \sim 50 \%$. 
This means it has a chance that the robot does not act as humans intended. 
To alleviate seamless brain-robot interaction, we introduce a "neural feedback" mechanism as illustrated in figure~\ref{fig:inference} part b). 
Here the human subject could observe whether the humanoid robot fits the intention, if not, the human subject could actively change the mental load or push more actively thinking and try multiple times until the neural signal successfully drives the robot. 
This novel interaction method could further let human subjects also better adapt to and drive the brain-humanoid-robot control system. 
\section{Experiment $\&$ Discussion}


 \begin{figure*}[hbpt]
    \centering
    \includegraphics[width=1\linewidth]{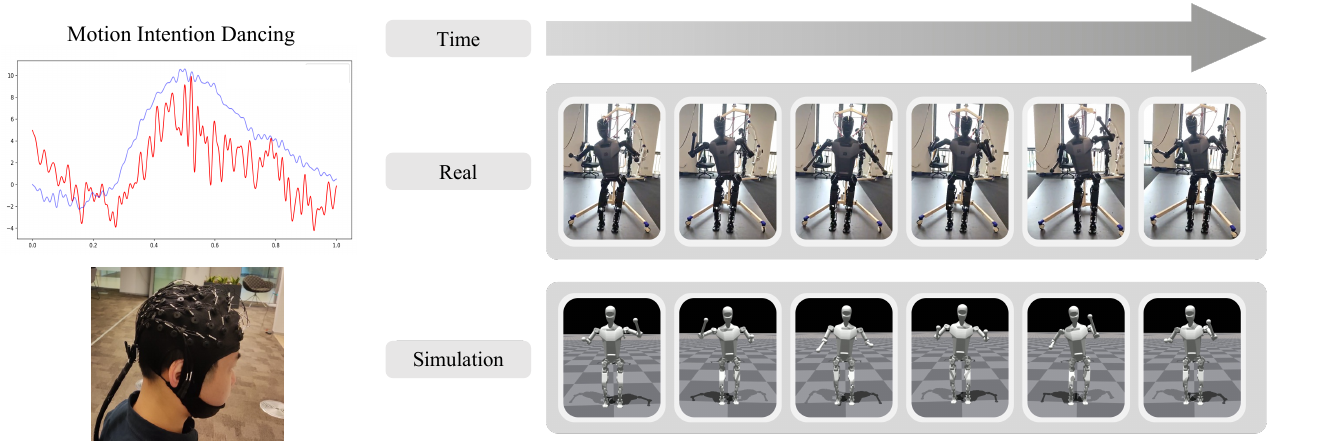}
    \caption{Illustration of the join brain-to-robotic control, where the left side shows the ERP curve when the human subject has motion intention "dance". The right side denotes the simulation motion trajectory generated and the real robot action according to these trajectories.   \label{fig:jointcontrol} }
\end{figure*}

\subsection{Robotic Training}
\label{subsec:joint}
We utilize the optical motion capture dataset AMASS~\cite{mahmood2019amass} and combine it with real data collected using the Xsens~\endnote{\href{https://www.movella.com/products/xsens}{https://www.movella.com/products/xsens}} motion capture device, training with a mixture of these two types of data. AMASS is a large-scale human motion database that unifies different optical marker motion capture datasets under a common framework and parameterization. It contains over 40 hours of motion data, covering more than 300 subjects and over 11,000 motions. AMASS uses the SMPL human model, a generative human model based on a blend of shape and pose space, capable of describing human shape and pose with a small number of parameters. AMASS employs a new method called MoSh++, which converts motion capture data into realistic 3D human meshes represented by rigid body models. This method is applicable to any marker set and can also recover soft tissue dynamics and realistic hand movements. The AMASS dataset is rigorously screened and quality-controlled, providing a rich and diverse set of human motion samples that can be used to train and evaluate methods for human pose and shape estimation.

Due to the differences in limb length, joint degrees of freedom, and other factors between humanoid robots and the datasets, as well as the variations in height and limb dimensions among the motion capture actors, it is necessary to perform kinematic remapping of these high-quality human motion data. By integrating the kinematic desired trajectories with the humanoid robot Tien Kung, we can use these trajectories as reward signals during the reinforcement learning training process. This approach enables the precise control of the desired positions for each joint of the robot.
These details are consistent with previous research on full-body control of humanoid robots~\cite{fu2024humanplus,zhang2024wococo,cheng2024expressive,zhang2024whole}.

\subsection{Brain to Motion Intention}
\textbf{Data Statistics}
The data collection duration consists of 2 months across 10 human subjects. The statistical results of the collected data are shown in table~\ref{tb:std}, where the data contains 1500 trials and 2.36 hours for each subject. The dataset contains a total of 23.6 hours of EEG data. 
\begin{table}[hbpt]
\centering
\caption{Statistical results on the collected EEG-motion synchronized data.~\label{tb:std}}
\begin{tabular}{c|c}
\toprule
\textbf{Metric}                                  & \textbf{Count}                               \\ \hline
Recorded Subjects                                & 10                                           \\ \midrule
Experiment Blocks                                & 4                                            \\ \midrule
Experiment Sessions                              & 16                                           \\ \midrule
Vocabulary Size                                  & 24                                           \\ \midrule
Repetitions per word                             & 250                                          \\ \midrule
Trials per Subject                               & 6000                           \\ \bottomrule
\end{tabular}
\end{table}

We train a Conformer encoder with 2 Conformer blocks. We set the embedding dimension to 512 with 8 attention heads and the feed-forward dimension size to 1024. During training, we set the coefficient for the training loss as $\alpha = 0.5$ and $\beta = 0.5$. We optimize the parameters of the Conformer models using the AdamW optimizer with an initial learning rate of $1e^{-4}$ and a weight decay of 0.05. The learning rate warms up over the first 500 steps to $1e^{-2}$ and linearly decays to $1e^{-6}$. In all experiments, we set the batch size to 256 and train the model for 100 epochs. Training is performed on a 2$\times$A40 GPU with 48 GB of memory. 
The EEG model performance is reported in table~\ref{tb:eegmodel}.

\begin{figure}[hbpt]
    \centering
    \begin{subfigure}[b]{0.45\linewidth}
        \centering
        \includegraphics[width=1\linewidth]{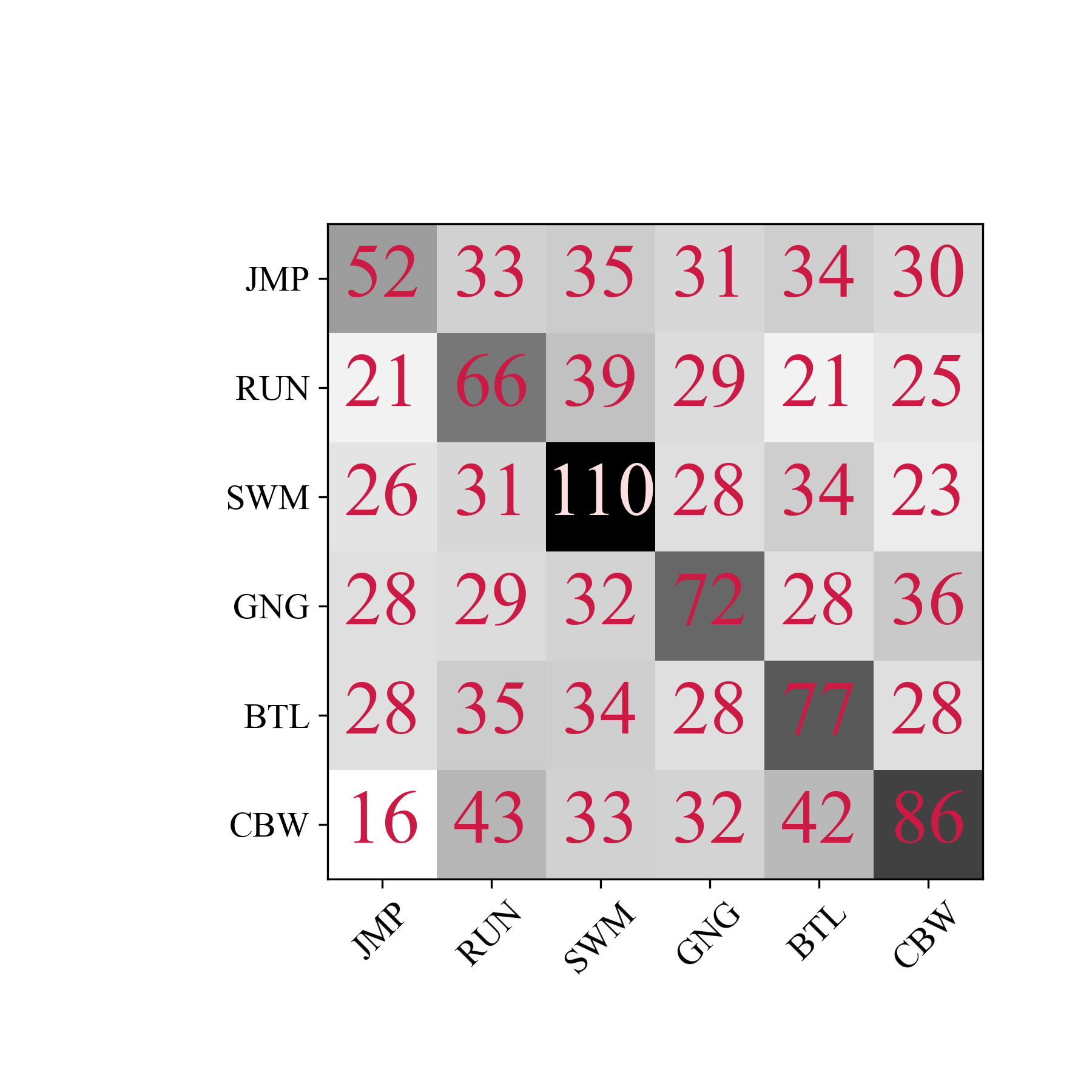}
        \caption{Confusion matrix within 6 classes.}
        \label{fig:confusion1}
    \end{subfigure}
    \begin{subfigure}[b]{0.45\linewidth}
        \centering
        \includegraphics[width=1\linewidth]{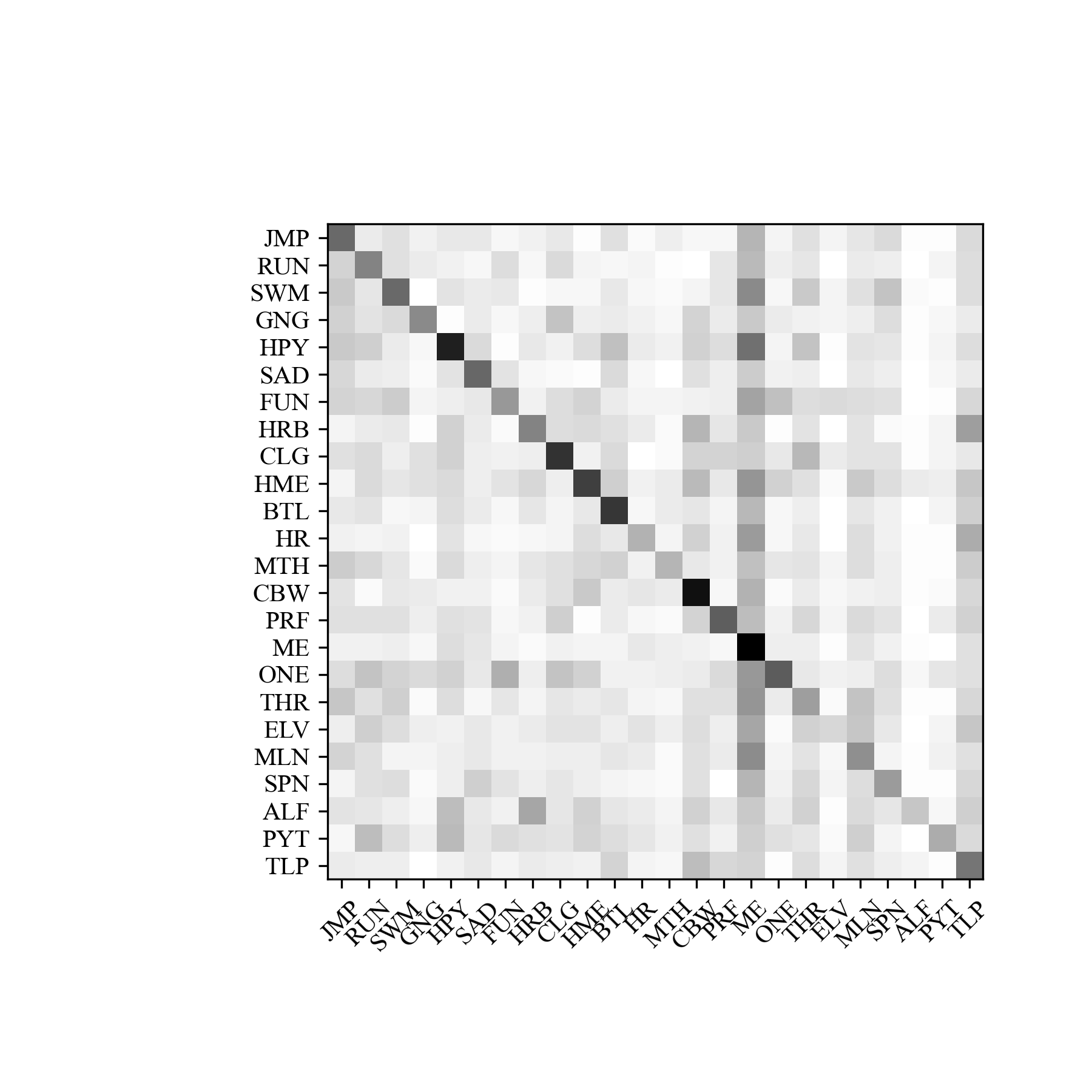}
        \caption{Confusion matrix within 24 classes.}
        \label{fig:confusion2}
    \end{subfigure}
    \caption{Confusion matrices of the classification results for two models.}
    \label{fig:confusion}
\end{figure}
Here, we evaluate the model accuracy with both the full set and the small common set (containing "running", "jumping", "dancing", "battle", and "cowboy").
We obtained an accuracy of 28.25\% top-1 accuracy for 24 classes while having an accuracy of 41.05\% top-1 accuracy for 6 motion-related classes. 
For top-3 accuracy, we could achieve 44.08\% and 64.21\% respectively on these two sets. 
To give a better understanding of the performance we provide the confusion matrix in figure~\ref{fig:confusion}.

\subsection{Brain to Robotic Control}

We illustrate the brain to robotic control demo by showing the ERP of the motion intention with the real demo as shown in figure~\ref{fig:jointcontrol}. 
Here to better show the difference between different motion intentions with ERP on the left side. The right part shows the motion trajectory generated from the motion intention (denoted as the simulation row) and the performance of the real robot (denoted as a real row).

In the joint inference stage, we test the \textbf{success rate} of the whole system by counting the ratio of the human subjects successfully driving the robot.
During the test period, the human subjects are given one motion intention and 3 minutes to try to drive the robot by multiple attempts with the neural adjustment through visual feedback (neural feedback). 
Four human subjects participated in the real-time demo test phase. The average \textbf{success rate} within 3 minutes is 52.34$\%$ for 6 class systems and 38.22$\%$ for 24 class systems. 


\section{Conclusion}

In this paper, we propose a novel framework, E2H, designed to enable brain-to-humanoid robot control. Due to the challenges of direct spatial decoding from low-quality EEG signals, the framework employs a two-stage process. In the first stage, brain signals are translated into semantic motion keywords. In the second stage, these keywords are converted into trajectory references, which are used by the robot control policy to guide the robot along the desired trajectory.
To develop and validate this approach, we collected 23.6 hours of synchronized EEG-to-motion data from 10 subjects. Our experimental results demonstrate that the brain decoding model achieves top-1 accuracies of 28.25$\%$ and 41.05$\%$ for 24-class and 6-class settings, respectively. Additionally, joint testing shows that the E2H framework enables successful control of physical robots after multiple attempts.
While the control accuracy remains limited at this stage, this research paves the way for future developments. Neural signal-driven humanoid robot control introduces new possibilities for human-robot interaction, offering significant potential for future applications.

\bibliographystyle{ACM-Reference-Format}
\bibliography{arxiv}
\theendnotes
\end{document}